\documentclass{article}
\usepackage[a4paper, total={6in, 8in}]{geometry}
\usepackage{fancyhdr}

\usepackage{listings}
\usepackage{bbm}
\usepackage{url}
\usepackage{booktabs}
\usepackage{algorithm}
\usepackage{xspace}
\usepackage[table]{xcolor}
\usepackage{subcaption}
\usepackage{amsmath,amssymb,amsfonts}

\usepackage{graphicx}
\usepackage{textcomp}
\usepackage{fancyvrb}
\usepackage{booktabs}
\usepackage{comment}
\usepackage{nicefrac}       
\usepackage{microtype}      
\usepackage{amsthm}
\usepackage{comment}

\usepackage{algpseudocode}

\usepackage{caption}
\usepackage{subcaption}
\usepackage{graphicx}
\graphicspath{{./figures/}}
\usepackage{tikz}
\usetikzlibrary{shapes, shadows,positioning}

\usepackage{multirow}
\usepackage{tabularx}
\usepackage{pgfplots}
\pgfkeys{/pgf/number format/.cd,1000 sep={}}

\newcommand{\paragraphb}[1]{\vspace{0.03in} \noindent{\bf #1} }

\newtheorem{theorem}{Theorem}

\newtheorem{lemma}{Lemma}
\newtheorem{definition}{Definition}

\newlength\figureheight
\newlength\figurewidth
\newlength\dualfigureheight
\newlength\dualfigurewidth
\newlength\trifigureheight
\newlength\trifigurewidth
\usepackage{harpoon}

\newlength\quadfigureheight
\newlength\quadfigurewidth
\newcommand{\expected}[2]{\underset{#2}{\mathbb{E}}[#1]}
\newcommand{\loss}{l}

\newcommand{\rdp}[0]{R\'enyi\xspace}

\usepackage{colortbl}

\usepackage{scalerel,stackengine}

\stackMath
\newcommand\widecheck[1]{%
\savestack{\tmpbox}{\stretchto{%
  \scaleto{%
    \scalerel*[\widthof{\ensuremath{#1}}]{\kern-.6pt\bigwedge\kern-.6pt}%
    {\rule[-\textheight/2]{1ex}{\textheight}}
  }{\textheight}%
}{0.5ex}}%
\stackon[1pt]{#1}{\scalebox{-1}{\tmpbox}}%
}

\algnewcommand{\Inputs}[1]{%
  \State \textbf{Inputs:}
  \Statex \hspace*{\algorithmicindent}\parbox[t]{.8\linewidth}{\raggedright #1}
}
\algnewcommand{\Outputs}[1]{%
  \State \textbf{Outputs:}
  \Statex \hspace*{\algorithmicindent}\parbox[t]{.8\linewidth}{\raggedright #1}
}
\algnewcommand{\Initialize}[1]{%
  \State \textbf{Initialize:}
  \Statex \hspace*{\algorithmicindent}\parbox[t]{.8\linewidth}{\raggedright #1}
}

\definecolor{lightgray}{gray}{0.9}

\date{}

\begin{document}

%
\title{Improving Deep Learning with Differential Privacy using Gradient Encoding and Denoising}

\author{
\textbf{Milad Nasr} \\
  UMass Amherst\\
  \texttt{milad@cs.umass.edu} \\
  \textbf{Reza Shokri} \\
  National University of Singapore\\
  \texttt{reza@comp.nus.edu.sg} \\
\textbf{Amir Houmansadr} \\
 UMass Amherst\\
 \texttt{amir@cs.umass.edu}
}

\maketitle

%
%

%

%

%
%
%

%
\begin{abstract}

 Deep learning models leak  significant amounts of information about their training datasets.  Previous work has investigated training models with differential privacy (DP) guarantees through adding DP noise to the gradients. However, such solutions (specifically, DPSGD), result in large degradations in the  accuracy of the trained models. In this paper, we aim at training deep learning models with DP guarantees while preserving  model accuracy much better than previous works.  Our key technique is to encode gradients to map them to a smaller vector space, therefore enabling us to obtain DP guarantees for different noise distributions. This allows us to investigate and choose  noise distributions that best  preserve model accuracy for a target privacy budget. We also take advantage of the post-processing property of differential privacy by introducing the idea of denoising,  which further improves the utility of the trained models without degrading their DP guarantees.  We show that our mechanism outperforms the state-of-the-art DPSGD; for instance, for the same model accuracy of $96.1\%$ on MNIST, our technique results in a privacy bound  of $\epsilon=3.2$ compared to $\epsilon=6$ of DPSGD, which is a significant improvement.

\end{abstract}



\section{Introduction}

Deep neural networks (DNN) are used in a wide range of learning applications. 
Unfortunately, the high capacities of DNNs make them susceptible to leaking private information about the datasets they use for training.
Recent works have demonstrated various types of privacy leakage in DNNs, most notably,  through membership inference~\cite{shokri2015privacy,nasrcomprehensive,carlini2018secret,melis2019exploiting} and model inversion~\cite{fredrikson2015model} attacks. 

A promising approach to alleviate such privacy leakage is to  train DNN models with  differential privacy (DP) guarantees. 
In particular, DPSGD by Abadi et al.~\cite{abadi2016deep} adds DP noise to clipped gradients, during the training process,  to train models with DP privacy protection. 
Unfortunately, existing DP-based solutions significantly reduce the utility (prediction accuracy) of the trained models~\cite{jayaraman2019evaluating}. 

In this work, we present a framework to train DNN models with DP guarantees that offer  much better tradeoffs between utility and privacy (i.e., they offer better prediction accuracies for the same privacy budgets).  
We use two key techniques in building our framework. 
First, while DPSGD uses Gaussian noise, recent works have shown that other noise distributions can improve utility for the same privacy budgets in different settings.
For instance,  Bun et al.~\cite{bun2019average} have shown that the Student-t noise  distribution can achieve better utility for trimmed mean applications. 
However, deriving the privacy bounds of DNN for arbitrary noise distributions is \emph{not} a trivial task; 
this is  
because estimating the privacy bound requires one to compute the distance between two probability distributions with infinite points, which is computationally infeasible and theoretically hard.
To be able to use arbitrary noise distributions, we \emph{encode} gradients into a finite vector space, and use numerical techniques to obtain DP privacy bounds for the given noise distribution. 
Specifically, we derive  privacy bounds by calculating the \rdp distances within our finite set of encoded gradients. 
This allows us to search in the space of various noise distributions to identify those with better utility-privacy tradeoffs.  

Our second key  technique is a \emph{denoising} mechanism that leverages the post-processing property of DP. Specifically, our denoising mechanism modifies privatized (i.e., noisy) gradients (by scaling them based on their useful information) to improve their contributions to  model's utility without reducing their DP privacy bounds.

We evaluate our framework on the CIFAR and MNIST  datasets, demonstrating that it outperforms DPSGD on the privacy-utility tradeoff. 
For instance, we show that  on MNIST, DPSGD achieves a $96.1\%$ prediction accuracy with a privacy budget $\epsilon=6$ while our framework achieves similar accuracy with an $\epsilon=3.2$ (using $\delta=10^{-5}$). Note that this is a \textbf{significant improvement} given the exponential impact of the privacy budget parameter ($\epsilon$) on privacy leakage.

\section{Background}

We overview the basics of  differential privacy and  deep learning. 
Table~\ref{tab:notations} lists the notations used across the paper. 

\begin{table*}[ht]
\caption{Description of notations}
\label{tab:notations}
\begin{center}
\rowcolors{1}{}{lightgray}
\begin{tabular}{r|l}
  \toprule
 Notation & Description  \\
   \midrule
 $D,D'$& training datasets \\
 $\mathcal{M}$ & privacy mechanism \\
 $\theta$ & model paramteres \\
 $\nabla_{\theta}$ & gradient w.r.t model parameters $\theta$ \\
 $\nabla_{\theta}^{\mu}$& gradient w.r.t model parameters $\theta$ for a  micro-batch \\
 $\widecheck{\nabla}_{\theta}^{\mu}$& encoded gradient w.r.t model parameters $\theta$ for a  micro-batch \\
 ${\nabla}_{\theta}^{G}$& aggregated vector of encoded gradients  for all  micro-batch in a minibatch\\
 $\widetilde{\nabla}_{\theta}^{G}$& privatized gradient (after adding noise) for a mini-batch \\
 $\widehat{\nabla}_{\theta}^{G}$& denoised (rescaled) privatized gradient vector for a mini-batch \\
 $\vec{\psi}$ & a preselected gradient vector for encoding \\
 $\Psi$ & the set of all preselected gradient vectors \\
 $Z_t$ & probability distribution for the private mechanisms\\
 $\mathcal{Z}$ &  the set of all probability distributions used in the learning \\ 
 $\epsilon, \delta$ & privacy bound metrics\\
   \bottomrule
\end{tabular}
\end{center}
\end{table*}

\subsection{Differential Privacy}
Differential privacy~\cite{dwork2011differential,dwork2014algorithmic} is the gold standard for data privacy.
It is formally defined as below:

\begin{definition}[{{Differential Privacy}}]
    A randomized mechanism  $\mathcal{M}$ with domain $\mathcal{D}$ and range $\mathcal{R}$ preserves $(\epsilon,\delta)-$differential privacy iff for any two neighboring   datasets $D,D' \in \mathcal{D}$ and for any subset $S \subseteq \mathcal{R}$ we have:
    \begin{align}
        \Pr[\mathcal{M}(D) \in S] \leq e^{\epsilon} \Pr[\mathcal{M}(D') \in S] + \delta
    \end{align}\label{def:dp}
   where $\epsilon$ is the \emph{privacy budget} and $\delta$ is the \emph{failure probability}. 
\end{definition}

R\'enyi Differential Privacy(RDP) is a commonly-used relaxed definition for differential privacy. 

\begin{definition}[{{R\'enyi Differential Privacy (RDP)}}~\cite{mironov2017renyi}]
 A randomized mechanism  $\mathcal{M}$ with domain $\mathcal{D}$ is $(\alpha,\epsilon)$-RDP with order $\alpha \in (1,\infty)$
iff for any two   neighboring datasets $D,D'\in \mathcal{D}$:
\begin{align}\label{eq:rdp}
	D_{\alpha} (\mathcal{M}(D)||\mathcal{M}(D')) := \frac{1}{\alpha - 1 } \log \expected{(\frac{\mathcal{M}(D)}{\mathcal{M}(D')})^{\alpha}}{\delta \sim \mathcal{M}(D')} \leq \epsilon
\end{align}

\end{definition}
\begin{lemma}[Adaptive Composition of RDP~\cite{mironov2017renyi}]\label{lem:composition}
	Consider two randomized mechanisms  $\mathcal{M}_1$ and $\mathcal{M}_1$ that provide  $(\alpha,\epsilon_1)$-RDP and  $(\alpha,\epsilon_2)$-RDP, respectively. Composing $\mathcal{M}_1$ and $\mathcal{M}_1$ results in a mechanism with 
 $(\alpha,\epsilon_1 + \epsilon_2)$-RDP.
\end{lemma}

\begin{lemma}[RDP to DP conversion~\cite{mironov2017renyi}]\label{lem:rdptodp}
	If $\mathcal{M}$ obeys $(\alpha,\epsilon)$-RDP, then $\mathcal{M}$  is $(\epsilon + \log(\frac{1}{\delta})/(\alpha-1), \delta)$-DP for all $\delta \in (0,1)$. 
\end{lemma}


\begin{lemma}[Post-processing of RDP~\cite{mironov2017renyi}]
Given a randomized mechanism  that is $(\alpha,\delta)-$\rdp differentially private, applying a randomized mapping function on it does not increase its  privacy budget, i.e., it will result in another $(\alpha,\delta)-$ \rdp differentially private mechanism.
\label{lemma:ec_post}
\end{lemma}

%

\subsection{Deep Learning with Differential Privacy}\label{ssec:dpsgd}


Several works have  used differential privacy in  traditional  machine learning algorithms to protect the privacy of the training data~\cite{li2014privacy,chaudhuri2011differentially,feldman2018privacy,zhang2016differential,bassily2014private}. 
Many of these works~\cite{feldman2018privacy, bassily2014private,chaudhuri2011differentially} use properties such as convexity or smoothness for their privacy analysis, which is not necessarily true in deep learning, and therefore, one cannot use many of such methods in practice.
Recently, Abadi et. al.~\cite{abadi2016deep} designed a deep learning training algorithm, DPSGD, 
where they used gradient clipping to limit the sensitivity of the learning algorithm, 
and then add noise to a clipped model gradient proportional to its sensitivity. 
They also introduced the \emph{momentum accountant} technique allowing them to  compute much tighter privacy bounds. 
In particular, 
they used the momentum accountant method to obtain high model predication accuracies using single digit privacy budgets. 
McMahan et al.~\cite{mcmahan2018learning} applied the momentum accountant method on language models. They also showed the feasibility of  using differential privacy in federated learning to achieve user-level privacy guarantees with an acceptable loss in model prediction accuracy.
Wang et al.~\cite{wang2018subsampled} showed that \rdp differential privacy can be used instead of accountant momentum to improve the privacy bounds. We use \rdp to compute the privacy bounds in this work. 
For a fair comparison, we also use \rdp differential privacy to compute the bounds for DPSGD.



\section{Our Framework}


In this paper, we present a \emph{generic framework} to learn models with differential privacy guarantees, expanding over prior works~\cite{abadi2016deep,mcmahan2018learning}.  
Our work differs from prior works in two ways: 
First, our framework allows us to use 
different noise distributions (ones that  have closed-form formulas and well-defined mean).
Second, we leverage the post processing property of differential privacy to enhance the learning procedure and improve  model accuracy without impacting privacy.

\paragraphb{Overview:} Here we  present an overview of our framework. 
In each iteration of training, 
we select a random mini-batch  from the training dataset.
We further divide the mini-batch into micro-batches, where the number of micro-batches is a hyperparameter. 
For each micro-batch, 
 we compute the gradient of the loss function ($\loss$) with respect to the  model parameters.
%
Then, we apply a particular \emph{encoding} on the obtained gradients, as will be explained. 
The purpose of this encoding is to map gradients into a finite vector space, allowing us to apply arbitrary noise distributions on the encoded gradients while being able to  compute their DP  privacy guarantees. 
Next, we aggregate  the encoded gradients from different micro-batches and apply a calibrated DP noise on them. 
Our framework allows us to use arbitrary noise distributions (thanks to our encoder); we evaluate the effect of different noise distributions on the performance of training. 
The final stage of our algorithm is a post-processing phase, called \emph{denoising}, in which  we use an error correction algorithm to improve the utility of the trained model given the adversary's prior knowledge.  The final gradients are then used to update the model. 



\begin{algorithm}[t!]
    \caption{Differentially Private Discrete SGD}
    \begin{algorithmic}[1]
    \Require{ learning rate $\eta$, $\mu$-batchsize $\mu$, batchsize $n$, noise models $\mathcal{Z}=\{Z_1,Z_2,\cdots,Z_T\}$, preselected gradient vectors $\Psi$}
    \State Initiate $\theta$ randomly
    \For{$t \in \{T\}$} 
    \State $B_t \gets$ Sample $n$ instances from dataset randomly 
    \State $\nabla_\theta^G \gets \vec{0}$ 
    \For{$\mu$-batch $b$  $\in B_t$}
    \State  $\nabla_\theta^{\mu} \gets$ gradient of micro-batch     $b$
    \State $\widecheck{\nabla_\theta^{\mu}} \gets $ encode gradients by solving equation \eqref{eq:select} 
    \Comment{Encoding}
    \State $\nabla_\theta^G \gets \nabla_\theta^G  + \widecheck{\nabla_\theta^{\mu} }$ 
    \EndFor
    \State $\mathcal{Z} \sim Z_t $
    \State $\widetilde{\nabla_\theta^G} \gets \nabla_\theta^G  + \mathcal{Z}$ \Comment{Noise Addition}
    \State $\widehat{\nabla_\theta^G} \gets$Error correction on $\widetilde{\nabla_\theta^G}$ \Comment{Error Correction}
    \State $\theta \gets \theta  - \eta \widehat{\nabla_\theta^G}$
    \EndFor

    \noindent
    \Return output $\theta$ 
    \end{algorithmic}
    \label{alg:main_alg}                       
\end{algorithm}

Below, we describe the two key components of our framework, gradient encoding and denoising. We will then present our privacy analysis.

\subsection{Gradient Encoding}\label{sec:coding}
\begin{figure*}
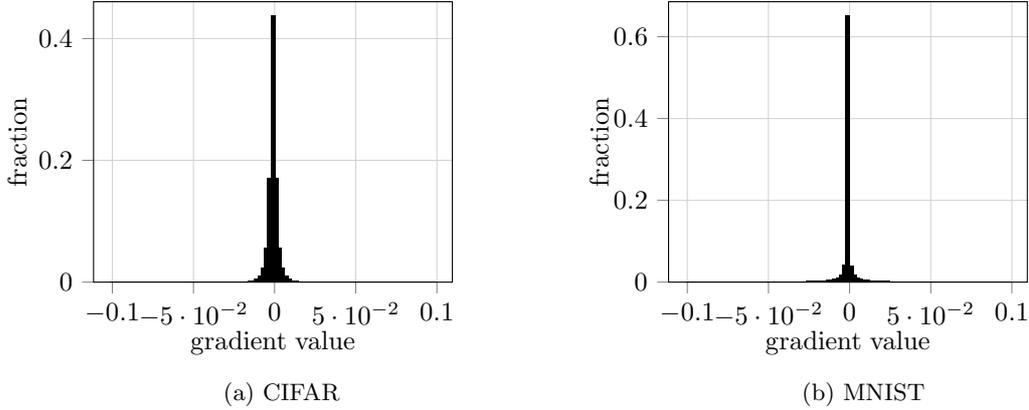

\centering
\begin{subfigure}[b]{0.49\linewidth}
	\input{figures/cifar_epoch_0}
	\caption{CIFAR}
\end{subfigure}
\hfill
\begin{subfigure}[b]{0.49\linewidth}
	\input{figures/mnist_epoch_0}
	\caption{MNIST}
\end{subfigure}
\hfill
    \caption{Distribution of gradient values  over all model parameters and data points at epoch 5
    }
    \label{fig:grad_distribution}
\end{figure*}

Deriving differential privacy guarantees  for arbitrary noise distributions is non-trivial; therefore, prior DP approaches for ML privacy~\cite{abadi2016deep,mcmahan2018learning} mainly stick to specific distributions like Gaussian distribution for which the privacy bounds can be computed. 
A promising approach to compute the differential privacy bounds of  a given noise distribution is to compute the \rdp distance (see \eqref{eq:rdp}) for all possible neighboring datasets. While we can analytically compute the bound for a few mechanisms (e.g. Gaussian~\cite{wang2018subsampled}),
this is not computationally feasible for other distributions unless  the input space is finite. 
To this aim, we use gradient encoding to map gradients into a finite space, enabling us to obtain \rdp DP bounds for arbitrary noise distributions. 

\paragraphb{Empirical distribution of  gradients:} In order to design our encoder, we first need to obtain the distribution of typical parameter gradients. We derive this distribution empirically by calculating the gradients for different models and datasets, and across different epochs of training.  
 Figure~\ref{fig:grad_distribution} shows the empirical distribution of  gradients for CIFAR and MNIST datasets in the first epoch of training, which can be fitted to a Gaussian distribution (the type of fitted distribution is not important in privacy bounds ).
%

\paragraphb{Sampling the distribution for mapping:} 
We sample the empirical distribution of gradients to obtain a set 
 $\Psi=\{\vec{\psi_1},\vec{\psi_2},\cdots,\vec{\psi_n}\}$ 
 where each $\vec{\psi_i}$ is a \emph{preselected gradient vector}, and $n$ is the number of preselected vectors.
 Each preselected gradient vector  $\vec{\psi_i}$ is a vector of size $k$, where $k$ is the size of the  model. For each $\vec{\psi_i}$, we draw its elements  i.i.d.\ from the distribution of gradients.

\paragraphb{Encoding gradients using preselected vectors:} To encode a gradient vector $\nabla_{\theta}$, we  replace it with a  preselected gradient vector  from $\Psi$ that is closest to $\nabla_{\theta}$; therefore, we  solve the following optimization problem:
\begin{align} \label{eq:select}
    Encode(\nabla_{\theta}) = \arg \min_{\vec{\psi_i} \in \Psi} ||\vec{\psi_i}- \nabla_{\theta}|| 
\end{align}
where $||\cdot||$ is the cosine distance. 
Below, we optimize the computation complexity of this mechanism. 	

\paragraphb{Practical speedup of encoding:}
Solving \eqref{eq:select} can be time/memory exhausting, since the size of each vector $\vec{\psi_i}$ equals   the size of the DNN model (typically, millions of parameters), 
and to perform an effective encoding, we need to have thousands of preselected gradient vectors (i.e., $\vec{\psi_i}$'s). We use the following adjustments to speed up solving \eqref{eq:select}.

First, 
based on the empirical distribution of gradients (Figure~\ref{fig:grad_distribution}),  
 we see that many  gradient values are close to zero; therefore, gradient vectors can be  represented using \emph{sparse} vectors.  
This allows us to reduce the size of each preselected gradient vector by a sparse representation,  which will  reduce both memory and time complexity of solving \eqref{eq:select}.

Second, using  the translation invariance property of RDP~\cite{mironov2017renyi}
we are able to order the elements of each gradient vector. 
We sort the gradient vectors before encoding, therefore, we only compute the distance between the sorted gradient vectors and sorted preselected gradient vectors, not every possible permutation.
Algorithm~\ref{alg:encodeing} summarizes our optimized mechanism for gradient encoding.

\begin{algorithm}[t!]
	\caption{Encoding gradients 
	}
    \begin{algorithmic}[1]
    \Require{Preselected sorted gradient vectors $\Psi$ and a  gradient vector $\nabla_\theta^{\mu}$ }
	    \State $\text{sorted-inds} \gets \text{argsort } |\nabla_\theta^{\mu}|$ 
	    \State $\vec{\psi} \gets \arg \min_{\vec{\psi} \in \Psi} \frac{\vec{\psi} . \nabla_\theta^{\mu}[sorted-inds] }{|\vec{\psi}| \times |\nabla_\theta^{\mu}|}$
	    \State $t \gets 0$
	    \State $\widecheck{\nabla_\theta^{\mu}} \gets \vec{0}$
	    \For{$i$ in sorted-indexes}
	    	\State $\widecheck{\nabla_\theta^{\mu}}[i]= \min(\vec{\psi}[t],\max(\nabla_\theta^{\mu}[i],-\vec{\psi}[t]))$ 
	    	\State $t = t+1$
	    \EndFor
    
    \noindent
    \Return $\widecheck{\nabla_\theta^{\mu}}$ \end{algorithmic}
	\label{alg:encodeing}                       
\end{algorithm}


\subsection{Denoising}

As overviewed before, encoded gradients from different micro-batches are aggregated and summed with a noise vector; we call the resulted vector a \emph{privatized gradient vector}.
We leverage the post-processing property of RDP (Lemma~\ref{lemma:ec_post}) 
to  modify  privatized gradient vectors in order to improve model utility (prediction accuracy) without impacting their RDP privacy bounds.

Our denoising process  \emph{scales} privatized gradients, based on their value to model utility, before using them to update the model. 
Consider an  oracle $O$ that  measures the ``usefullness'' (i.e., information value) of  a privatized gradient vector $\widetilde{\nabla_\theta^G}$ as $O(\widetilde{\nabla_\theta^G})$.
At the end of each iteration, our model is updated as $\theta \gets \theta  - \eta O(\widetilde{\nabla_\theta^G})\cdot\widetilde{\nabla_\theta^G}$. Note that based on the post-processing property of RDP, such denoising does not change RDP's privacy guarantees as long as $O$ is a randomized process which does not use the private data (Lemma~\ref{lemma:ec_post}).

We formulate the utility usefulness of a privatized gradient vector as the closeness (inverse distance) between that  privatized vector and its corresponding  original gradient vector (the vector before  noise addition). This is because, intuitively, the closer a private gradient vector is to the original gradient, it will contribute more to  model accuracy. 
However, using the original vector to compute the scaling factor (the distance) will effect RDP's privacy guarantees  as it will make use of the private data.
Instead, we approximate the usefulness of a privatized gradient with \emph{its distance from the noise distribution used by our algorithm} (note that we measure the distance with the noise ``distribution'', not the specific noise sample used by our training algorithm as doing the latter will violate privacy bounds). 
Therefore, we denoise a privatized gradient vector $\widetilde{\nabla_\theta^G}$ as $\text{KS}(\widetilde{\nabla_\theta^G},Z_t)\widetilde{\nabla_\theta^G}$, where $Z_t$ is the noise distribution and $\text{KS}(.)$ is the 
 Kolmogorov-Smirnov (KS) distance metric (note that we also experimented with other distance metrics,  and KS achieved the best results in our experiments). 
 Therefore, our denoising formula  is given by  $\widehat{\nabla_\theta^G} \gets \eta \text{KS}(\widetilde{\nabla_\theta^G},Z_t) \widetilde{\nabla_\theta^G}$.

\section{Privacy Analysis}

In this section, we analyze the privacy bounds of our private learning framework. 
We use R\'enyi differential privacy to evaluate RDP guarantees, which can be converted to   equivalent DP bounds using Lemma~\ref{lem:rdptodp}. 
To compute the RDP privacy parameter, we need to  bound the R\'enyi distance between any two neighboring training datasets $D,D' \in \mathcal{D}$, where  $\mathcal{D}$ is the underlying data distribution and $D' = D \cup \{n\}$. 
We start by computing the privacy bound for one iteration, then use the composition Lemma~\ref{lem:composition} to estimate an upper bound for the overall privacy parameter.

 Given a training dataset  $D$, our training process can be formulated as a  randomized algorithm  given by 
$\mathcal{M}(D)= \sum_{x \in b_i } \widecheck{\nabla}_{\theta}^x + \mathcal{Z}$, where $b_i$ is a mini-batch sampled from $D$, $\mathcal{Z}$ is the noise, and $\widecheck{\nabla}_{\theta}^x$ is encoded gradient of the micro-batch $x$ (we have omitted the denoising function as it does not impact privacy analysis). 
 Now, we need to solve the following problem to bound the R\'enyi distance:
%
\begin{align}
        \sup_{|D \backslash D'| =1 } D_{\alpha} (\mathcal{M}(D)||\mathcal{M}(D'))
 \end{align}

We know that each encoded gradient belongs to  a finite set of preselected  vectors, say a specific vector $\vec{\psi}\in \Psi$. 
Now we  find the \rdp differential privacy of one iteration given this specific gradient vector.

\begin{lemma}\label{lemma:basic}
Consider one iteration of Algorithm~\ref{alg:main_alg}, and suppose there is only one preselected gradient vector, i.e., $\Psi=\{\vec{\psi}\}$.
Algorithm~\ref{alg:main_alg}  with the sampling rate $q$, and a probability distribution described by its pdf $z(.;\mu)$, where $\mu$ is the mean of the distribution, obeys $(\alpha,\epsilon)-$RDP, for a given $\alpha \in \mathcal{N}\slash \{1\}$;  $\epsilon$ can be computed as follows:
\begin{align}
    &\epsilon(\alpha; q, \vec{\psi}, z) \leq \\ \nonumber
    & \frac{1}{1-\alpha} \log \sum_{k=0}^{\alpha} \binom{\alpha}{k} q^k (1-q)^{\alpha-k} \prod_{\tau \in  \vec{\psi}} \int_{-\infty}^{\infty} (\frac{z(x;\tau)}{z(x;0)})^k z(x;0) dx
\end{align}

\begin{proof}
Please refer to the supplementary material document.	
\end{proof}

\end{lemma}




Next, we extend  Lemma~\ref{lemma:basic} to the case where $\Psi$ has more that one element.  To do this, we compute the bound for each element of  $\Psi$ separately, and then compute the maximum bound to obtain an upper bound on the \rdp distance. Lemma~\ref{lemma:basic2} summarizes this.

\begin{lemma}\label{lemma:basic2}
    One iteration of Algorithm~\ref{alg:main_alg}  with a set of  preselected gradient  vectors $\Psi=\{\psi_1,\psi_2,\cdots,\psi_n\}$, a sampling rate $q$, and a probability distribution described by its pdf $z(.;\mu)$, where $\mu$ is the mean of the distribution, obeys $(\alpha,\epsilon)-$RDP for a given $\alpha \in \mathcal{N}\slash \{1\}$;  $\epsilon$ can be computed as follow:
\begin{align}
    &\epsilon(\alpha; q, \Psi , z) \leq \max_{\psi \in \Psi} \epsilon(\alpha; q, \psi , z)
\end{align}
\begin{proof}
	For any change in the training dataset $D$, the aggregated gradient will change by one preselected gradient vector, therefore, the distance will not change more than the maximum possible distance.
\end{proof}
\end{lemma}

Now, using Lemma~\ref{lemma:basic2}, we can compute the privacy bounds for one iteration of our learning framework. 
Finally,  we can use the composition theorem of RDP  to bound the final privacy leakage of our learning framework.

\begin{theorem}\label{thrm:all}
    Algorithm~\ref{alg:main_alg} with a  set of preselected gradient  vectors $\Psi=\{\psi_1,\psi_2,\cdots,\psi_n\}$,  sampling rate $q$, number of iterations $T$, and probability distributions $\mathcal{Z}=\{Z_1,Z_2,\cdots,Z_T\}$ described by their pdf $z_t(.;\mu)$, where $\mu$ is the mean of the distribution, obeys $(\alpha,\epsilon)-$RDP, for a given $\alpha \in \mathcal{N}\slash \{1\}$;  $\epsilon$ can be computed as follow:
\begin{align}
    &\epsilon(\alpha; q, \Psi,\mathcal{Z},T) \leq \sum_{t=1}^T \epsilon(\alpha; q, \Psi , z_t)
\end{align}
\begin{proof}
	Using Lemma~\ref{lemma:basic2} and the composition Lemma~\ref{lem:composition}.
\end{proof}
\end{theorem}

As  mentioned earlier, since the encoded gradients belong to a finite set $\Psi$, we can compute the privacy bound even for probability distributions where we do not know the general formulation of their \rdp distances.


\section{Experimental Results}
\label{sec:experiments}

\paragraphb{Setup.}
We have implemented our algorithm in PyTorch~\cite{paszke2017automatic} and Tensorflow~\cite{tensorflow}. Unlike SGD, we need to compute the gradients per sample (or per $\mu-$batch), which makes it computationally expensive. We leverage multiple GPUs in parallel to speed up the training process. 
We train models on MNIST and CIFAR10 datasets, with cross-entropy as the loss function.  We choose a training set up comparable to that of Abadi et al.~\cite{abadi2016deep}.\footnote{Available at~\url{https://github.com/tensorflow/privacy}}

\begin{figure}
\centering
	\begin{subfigure}[b]{0.5\textwidth}
\begin{tikzpicture}
\begin{axis}[
height=\dualfigureheight,
    width=\dualfigurewidth,
legend cell align={left},
legend style={fill opacity=0.8, draw opacity=1, text opacity=1, at={(0.95,0.03)}, anchor=south east, draw=white!80!black},
tick align=outside,
tick pos=left,
x grid style={white!69.0196078431373!black},
xlabel={Accuracy},
xmajorgrids,
xmin=54, xmax=58.7355,
xtick style={color=black},
xtick={54,55,56,57,58,59},
xticklabels={\(\displaystyle 54\),\(\displaystyle 55\),\(\displaystyle 56\),\(\displaystyle 57\),\(\displaystyle 58\),\(\displaystyle 59\)},
y grid style={white!69.0196078431373!black},
ylabel={\(\displaystyle \epsilon\)},
ymajorgrids,
ymin=2.4498144651203, ymax=7.32143740642284,
ytick style={color=black},
ytick={2,3,4,5,6,7,8},
yticklabels={\(\displaystyle 2\),\(\displaystyle 3\),\(\displaystyle 4\),\(\displaystyle 5\),\(\displaystyle 6\),\(\displaystyle 7\),\(\displaystyle 8\)}
]
\addplot [thick, black]
table {%
55.85 7.1
55.68 6.23
55.46 5.1
55.27 4.34
54.94 3.54
54.14 2.81
};
\addlegendentry{DPSGD~\cite{abadi2016deep}}
\addplot [thick, red, dashed]
table {%
56.3 7.1
56.1 6.23
55.97 5.1
55.47 4.34
54.38 2.9
54 2.81
};
\addlegendentry{Alg~\ref{alg:main_alg}- Gaussian }
\addplot [thick, blue, dotted]
table {%
58.51 6.61421637445521
57.99 5.82884370120998
57.75 5.24349280386186
56.83 4.75066463330779
56.7 4.36972135499457
56.52 4.02776027770498
55.78 3.50799968452107
55.24 3.30028024704303
54.81 3.11508230425542
54.4 2.67125187154314
};
\addlegendentry{Alg~\ref{alg:main_alg}- Student-t}
\end{axis}

\end{tikzpicture}
	\caption{CIFAR10}
	\end{subfigure}
	\begin{subfigure}[b]{0.48\textwidth}
\begin{tikzpicture}

\begin{axis}[
height=\dualfigureheight,
    width=\dualfigurewidth,
legend cell align={left},
legend style={fill opacity=0.8, draw opacity=1, text opacity=1, at={(0.03,0.97)}, anchor=north west, draw=white!80!black},
tick align=outside,
tick pos=left,
x grid style={white!69.0196078431373!black},
xlabel={Accuracy},
xmajorgrids,
xmin=93.773, xmax=97.887,
xtick style={color=black},
y grid style={white!69.0196078431373!black},
ylabel={\(\displaystyle \epsilon\)},
ymajorgrids,
ymin=1.50014026647087, ymax=8.53787541811942,
ytick style={color=black},
ytick={0,2,4,6,8,10},
yticklabels={1,2,3,4,5,6}
]
\addplot [thick, black]
table {%
95.96 8.21797836577176
95.6 4.92754555629587
95.5 4.34260818783893
95.3 3.54586053777188
95.1 3.00915534936541
94.8 2.47404363854457
94.41 1.92088824550378
};
\addlegendentry{DPSGD~\cite{abadi2016deep}}
\addplot [thick, red, dashed]
table {%
97.01 8.21797836577176
96.76 4.92754555629587
96.61 4.34260818783893
95.93 3.54586053777188
95.53 3.00915534936541
94.75 2.47404363854457
94.45 1.92088824550378
};
\addlegendentry{Alg~\ref{alg:main_alg}- Gaussian}
\addplot [thick, blue, dotted]
table {%
97.26 8.1
97.21 7
97.14 6.1
97.1 5.44941360903564
96.74 4.09059438793042
96.41 3.65477418142667
96.35 3.29592774820714
96.15 3.01218809697881
95.97 2.78090410561696
95.46 2.40858148981197
94.99 2.14246504190255
94.39 1.9562553456522
93.96 1.82003731881854
};
\addlegendentry{Alg~\ref{alg:main_alg}- Student-t}
\end{axis}

\end{tikzpicture}
	\caption{MNIST}
	\end{subfigure}
	\caption{Comparing the privacy bound of our framework with DPSGD~\cite{abadi2016deep} for different model accuracies. We use a constant number of iterations and a  $\delta=10^{-5}$.}
	\label{fig:fix_epochs}
\end{figure}
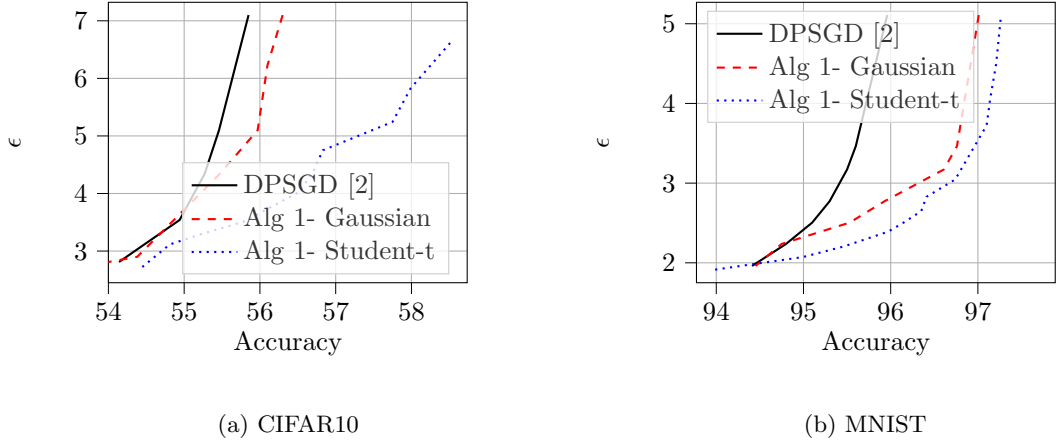

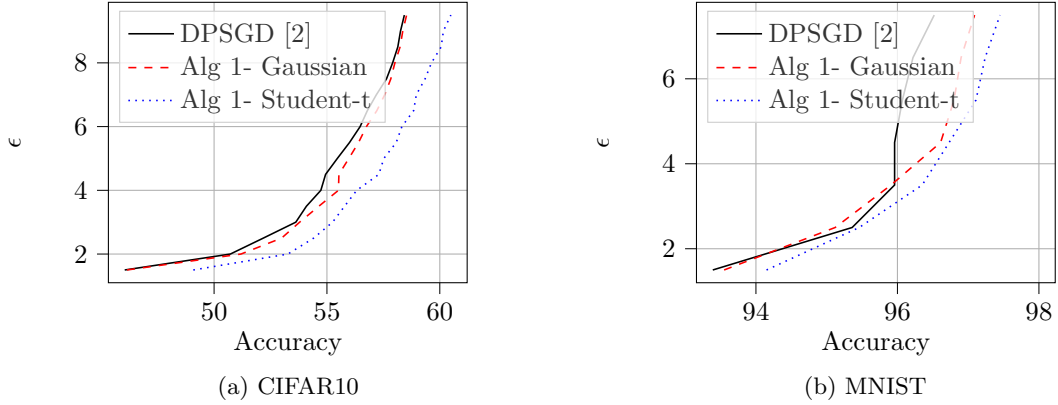
\begin{figure}
\centering
	\begin{subfigure}[b]{0.5\textwidth}
\begin{tikzpicture}

\begin{axis}[
height=\dualfigureheight,
    width=\dualfigurewidth,
legend cell align={left},
legend style={fill opacity=0.8, draw opacity=1, text opacity=1, at={(0.03,0.97)}, anchor=north west, draw=white!80!black},
tick align=outside,
tick pos=left,
x grid style={white!69.0196078431373!black},
xlabel={Accuracy},
xmajorgrids,
xmin=45.3175, xmax=61.2125,
xtick style={color=black},
y grid style={white!69.0196078431373!black},
ylabel={\(\displaystyle \epsilon\)},
ymajorgrids,
ymin=1.1, ymax=9.9,
ytick style={color=black}
]
\addplot [semithick, black]
table {%
46.04 1.5
50.72 2
52.18 2.5
53.62 3
54.09 3.5
54.73 4
54.94 4.5
55.46 5
56 5.5
56.47 6
56.79 6.5
57.19 7
57.64 7.5
57.91 8
58.14 8.5
58.26 9
58.43 9.5
};
\addlegendentry{DPSGD~\cite{abadi2016deep}}
\addplot [semithick, red, dashed]
table {%
46.14 1.5
51.19 2
52.99 2.5
53.79 3
54.64 3.5
55.51 4
55.53 4.5
55.97 5
56.4 5.5
56.75 6
57.21 6.5
57.56 7
57.83 7.5
58.01 8
58.24 8.5
58.36 9
58.53 9.5
};
\addlegendentry{Alg~\ref{alg:main_alg}- Gaussian}
\addplot [semithick, blue, dotted]
table {%
49.08 1.5
53.27 2
54.41 2.5
55.25 3
55.78 3.5
56.32 4
57.3 4.5
57.52 5
58.07 5.5
58.32 6
58.84 6.5
58.96 7
59.35 7.5
59.63 8
60.06 8.5
60.16 9
60.49 9.5
};
\addlegendentry{Alg~\ref{alg:main_alg}- Student-t}
\end{axis}

\end{tikzpicture}
	\vspace*{-0.5cm}
	\caption{CIFAR10}
	\end{subfigure}
	\begin{subfigure}[b]{0.48\textwidth}
\begin{tikzpicture}

\begin{axis}[
height=\dualfigureheight,
    width=\dualfigurewidth,
legend cell align={left},
legend style={fill opacity=0.8, draw opacity=1, text opacity=1, at={(0.03,0.97)}, anchor=north west, draw=white!80!black},
tick align=outside,
tick pos=left,
x grid style={white!69.0196078431373!black},
xlabel={Accuracy},
xmajorgrids,
xmin=93.1595, xmax=98.2305,
xtick style={color=black},
y grid style={white!69.0196078431373!black},
ylabel={\(\displaystyle \epsilon\)},
ymajorgrids,
ymin=1.2, ymax=7.8,
ytick style={color=black}
]
\addplot [semithick, black]
table {%
93.39 1.5
95.36 2.5
95.96 3.5
95.96 4.5
96.07 5.5
96.22 6.5
96.52 7.5
};
\addlegendentry{DPSGD~\cite{abadi2016deep}}
\addplot [semithick, red, dashed]
table {%
93.55 1.5
95.12 2.5
95.91 3.5
96.61 4.5
96.8 5.5
96.9 6.5
97.09 7.5
};
\addlegendentry{Alg~\ref{alg:main_alg}- Gaussian}
\addplot [semithick, blue, dotted]
table {%
94.15 1.5
95.46 2.5
96.35 3.5
96.73 4.5
97.11 5.5
97.24 6.5
97.45 7.5
};
\addlegendentry{Alg~\ref{alg:main_alg}- Student-t}
\end{axis}

\end{tikzpicture}
	\vspace*{-0.5cm}
	\caption{MNIST}
	\end{subfigure}
	\caption{Comparing \emph{the best} privacy bound of each mechanism for  different accuracy values (to find the best privacy bound, we change various hyper-parameters such as the number of epochs, sample rates, and noise parameters)}
	\label{fig:diff_epochs}
\end{figure}



In our experiments,  we used random sampling with replacement to create our training mini-batches (which is the correct way of creating the training batches based on the analysis).\footnote{This is  why we see some inconsistencies between our results and those at \url{https://github.com/tensorflow/privacy} which uses shuffling to create mini-batches.} 

\paragraphb{Comparing  performances for constant epochs.}
Figure~\ref{fig:fix_epochs} compares the accuracy-$\epsilon$ trade off of our system with the state-of-the-art system of DPSGP~\cite{abadi2016deep} for CIFAR and MNIST. As mentioned earlier, our framework allows using arbitrary noise distributions; we are only showing our results when we use Gaussian and Student-t distributions, which performed best among all the noise distributions we experimented (we also experimented with  Laplace, Cauchy, Variance Gamma (with $\lambda=2$), and Sech distributions). 
To have fair evaluations, we use the same number of epochs for our framework as well as DPSGD (100 for CIFAR and 60 for MNIST). 
To achieve different  privacy bounds, we varied the  parameters of each of the  noise distributions; for Gaussian, we used variances from 0.8 to 1.4, and for Student-t, we used 9 degrees of freedom with variances from 0.9 to 1.5.
Note that  we used a non-standardized form of Student-t distribution to be able to  change the variance independent of the freedom degree. 
We also used the same learning rate, model architecture, and hyper-parameters  across different noise distributions and methods. 
To select the preselected gradient vectors, we sample 1000 vectors with sizes similar to the size of the model gradient from a standard Gaussian distribution, and we replaced the values smaller than $10^{-5}$ with zero to avoid underflow in our privacy bound calculations. 
Then we clip the norm of each vector to be equal to one (to have a fair comparison with DPSGD). 
Also note that we use the standard $(\epsilon,\delta)$-DP metric for easier comparison with DPSGD (we convert RDP parameters to DP using Lemma~\ref{lem:rdptodp}).
In all experiments we set  $\delta=10^{-5}$ similar to DPSGD. 

 As shown in Figure~\ref{fig:fix_epochs}, our framework outperforms the state-of-the-art DPSG by large margins. Specifically,  using the Student-t distribution we can achieve similar accuracies with better privacy bounds (or equivalently,  better accuracy with similar privacy budget). For instance, we see that 
for CIFAR our framework achieves a $55.5$\% accuracy with $\epsilon=3.6$ compared to DPSGD's $ \epsilon=5$,   and on MNIST we  achieve a $96$ accuracy with $\epsilon=2.5$ where DPSGD reaches to same accuracy with  $\epsilon=5$.
We outperform DPSGD for two  main reasons. 
 First, framework allows using arbitrary noise distributions, and as can be seen, using Student-t noise results in  a better performance compared to Gaussian (which is used by DPSGD). 
 Second, we see that even our framework with Gaussian noise outperforms DPSGD (which also use Gaussian), which is due to our denoising mechanism. 
 

\paragraphb{Comparing the best performances.}
In our evaluations of Figure~\ref{fig:fix_epochs}, we used the same, constant number of iterations across the systems (100 for CIFAR and 60 for MNIST). 
Alternatively, in  Figure~\ref{fig:diff_epochs} we report the \emph{best} accuracy each system can achieve for a given privacy budget but by varying other settings. 
Specifically, for each privacy budget, we tried different combinations of noise model parameters ($Z$), number of iterations ($T$), and sampling rates $q$,  and reported the best results for each system. 
Note that, in practice searching the hyper-parameters for the best results will impact privacy bounds~\cite{abadi2016deep}, however, we do this experiment to compare the best performances of different systems. 
From Figure~\ref{fig:diff_epochs} we see that our system still outperforms DPSGD. 
For instance, for CIFAR our framework  achieves a $55\%$ accuracy with $\epsilon=3$ while DPSGD achieves similar accuracy  with $\epsilon=4.2$. Similarly, on MNIST we achieve a $96.1$\% accuracy with $\epsilon=3.2$ compared to $\epsilon=6$ for DPSGD.
This shows a \emph{significant improvement} given the exponential impact of $\epsilon$ on privacy leakage.
Also by comparing Figure~\ref{fig:fix_epochs} and Figure~\ref{fig:diff_epochs},
we see a  similar gap exists between Student-t and DPSGD. 

\paragraphb{Other takeaways.} An interesting  takeaway of our experiments is the \emph{exponential behavior of the privacy budget $\epsilon$ with respect to model accuracy}. 
Therefore, for higher accuracies, we need to spend much larger privacy budgets to further improve model accuracy. For instance, in the MNIST dataset, we can improve the accuracy by one percent from $94\%$ to $95\%$ by increasing the privacy budget by $0.6$, however, increasing the accuracy from $96\%$ to $97\%$ requires to increase the privacy budget  by $2$. 
We observe a similar behavior for CIFAR. 
Note that the  privacy budget itself has an exponential impact on privacy, which further demonstrates the difficulties of training private models with high accuracies. 

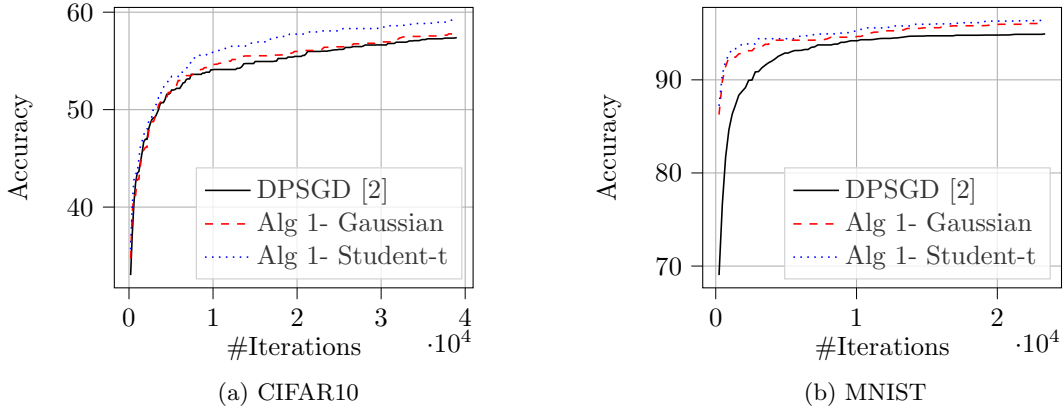
\begin{figure}[h]
\centering
	\begin{subfigure}[b]{0.5\textwidth}
\begin{tikzpicture}

\begin{axis}[
height=\dualfigureheight,
    width=\dualfigurewidth,
legend cell align={left},
legend style={fill opacity=0.8, draw opacity=1, text opacity=1, at={(0.97,0.03)}, anchor=south east, draw=white!80!black},
tick align=outside,
tick pos=left,
x grid style={white!69.0196078431373!black},
xlabel={\#Iterations},
xmajorgrids,
xmin=-1745.25, xmax=40940.25,
xtick style={color=black},
y grid style={white!69.0196078431373!black},
ylabel={Accuracy},
ymajorgrids,
ymin=31.712, ymax=60.488,
ytick style={color=black}
]
\addplot [semithick, black]
table {%
195 33.02
390 37.24
585 40.37
780 42.45
975 43.47
1170 43.64
1365 44.62
1560 45.59
1755 46.71
1950 46.96
2145 46.96
2340 48.02
2535 48.57
2730 49
2925 49.15
3120 49.25
3315 49.59
3510 49.99
3705 50.59
3900 50.72
4095 50.72
4290 51.03
4485 51.57
4680 51.57
4875 51.67
5070 52.01
5265 52.01
5460 52.06
5655 52.18
5850 52.18
6045 52.18
6240 52.46
6435 52.74
6630 52.79
6825 53.1
7020 53.16
7215 53.16
7410 53.62
7605 53.62
7800 53.62
7995 53.62
8190 53.62
8385 53.62
8580 53.62
8775 53.75
8970 53.75
9165 53.84
9360 53.84
9555 53.84
9750 54.09
9945 54.09
10140 54.12
10335 54.12
10530 54.12
10725 54.12
10920 54.12
11115 54.12
11310 54.12
11505 54.12
11700 54.12
11895 54.12
12090 54.12
12285 54.12
12480 54.12
12675 54.12
12870 54.24
13065 54.24
13260 54.29
13455 54.44
13650 54.7
13845 54.73
14040 54.73
14235 54.73
14430 54.73
14625 54.73
14820 54.73
15015 54.94
15210 54.94
15405 54.94
15600 54.94
15795 54.94
15990 54.94
16185 54.94
16380 54.94
16575 54.94
16770 54.94
16965 54.94
17160 54.94
17355 54.98
17550 54.98
17745 55.25
17940 55.25
18135 55.25
18330 55.25
18525 55.25
18720 55.25
18915 55.39
19110 55.39
19305 55.39
19500 55.39
19695 55.46
19890 55.46
20085 55.46
20280 55.46
20475 55.46
20670 55.56
20865 55.56
21060 55.72
21255 55.97
21450 55.97
21645 55.97
21840 55.97
22035 55.97
22230 55.97
22425 55.97
22620 55.97
22815 55.98
23010 56
23205 56
23400 56
23595 56
23790 56.05
23985 56.05
24180 56.11
24375 56.12
24570 56.14
24765 56.14
24960 56.18
25155 56.18
25350 56.18
25545 56.2
25740 56.38
25935 56.38
26130 56.38
26325 56.38
26520 56.43
26715 56.47
26910 56.47
27105 56.47
27300 56.47
27495 56.47
27690 56.47
27885 56.53
28080 56.63
28275 56.63
28470 56.63
28665 56.63
28860 56.63
29055 56.63
29250 56.63
29445 56.64
29640 56.64
29835 56.64
30030 56.64
30225 56.64
30420 56.64
30615 56.64
30810 56.79
31005 56.79
31200 56.79
31395 56.79
31590 56.8
31785 56.92
31980 56.92
32175 56.92
32370 56.92
32565 56.92
32760 56.92
32955 56.92
33150 56.97
33345 57.05
33540 57.05
33735 57.05
33930 57.08
34125 57.08
34320 57.08
34515 57.08
34710 57.14
34905 57.19
35100 57.19
35295 57.19
35490 57.27
35685 57.27
35880 57.27
36075 57.27
36270 57.27
36465 57.27
36660 57.27
36855 57.27
37050 57.27
37245 57.27
37440 57.31
37635 57.34
37830 57.34
38025 57.34
38220 57.34
38415 57.34
38610 57.34
38805 57.36
39000 57.39
};
\addlegendentry{DPSGD~\cite{abadi2016deep}}
\addplot [semithick, red, dashed]
table {%
195 34.71
390 39.48
585 41.2
780 41.2
975 42.84
1170 42.84
1365 44.32
1560 45.55
1755 45.84
1950 46.13
2145 46.14
2340 47.81
2535 48.2
2730 48.77
2925 48.77
3120 49.26
3315 49.71
3510 50.47
3705 50.73
3900 50.88
4095 51.19
4290 51.23
4485 51.23
4680 51.76
4875 51.76
5070 52.05
5265 52.15
5460 52.29
5655 52.49
5850 52.99
6045 52.99
6240 52.99
6435 53.25
6630 53.25
6825 53.49
7020 53.49
7215 53.49
7410 53.49
7605 53.64
7800 53.79
7995 53.79
8190 53.9
8385 54.04
8580 54.13
8775 54.13
8970 54.31
9165 54.31
9360 54.33
9555 54.6
9750 54.61
9945 54.64
10140 54.64
10335 54.66
10530 54.67
10725 54.76
10920 54.76
11115 54.85
11310 54.85
11505 54.87
11700 54.87
11895 54.97
12090 54.97
12285 55.15
12480 55.15
12675 55.15
12870 55.29
13065 55.29
13260 55.29
13455 55.43
13650 55.48
13845 55.51
14040 55.51
14235 55.51
14430 55.51
14625 55.51
14820 55.53
15015 55.53
15210 55.53
15405 55.53
15600 55.53
15795 55.53
15990 55.53
16185 55.53
16380 55.53
16575 55.6
16770 55.6
16965 55.6
17160 55.6
17355 55.6
17550 55.6
17745 55.6
17940 55.6
18135 55.62
18330 55.62
18525 55.62
18720 55.62
18915 55.78
19110 55.78
19305 55.78
19500 55.85
19695 55.97
19890 55.97
20085 55.97
20280 56.04
20475 56.07
20670 56.07
20865 56.07
21060 56.07
21255 56.07
21450 56.07
21645 56.07
21840 56.07
22035 56.07
22230 56.07
22425 56.07
22620 56.1
22815 56.1
23010 56.23
23205 56.4
23400 56.4
23595 56.4
23790 56.4
23985 56.4
24180 56.45
24375 56.45
24570 56.45
24765 56.45
24960 56.45
25155 56.45
25350 56.45
25545 56.45
25740 56.45
25935 56.45
26130 56.45
26325 56.62
26520 56.62
26715 56.75
26910 56.75
27105 56.75
27300 56.75
27495 56.75
27690 56.75
27885 56.81
28080 56.81
28275 56.81
28470 56.81
28665 56.81
28860 56.81
29055 56.81
29250 56.81
29445 56.95
29640 56.95
29835 56.95
30030 56.95
30225 56.95
30420 56.95
30615 56.95
30810 56.95
31005 56.95
31200 56.95
31395 57.21
31590 57.21
31785 57.21
31980 57.21
32175 57.35
32370 57.38
32565 57.46
32760 57.46
32955 57.47
33150 57.47
33345 57.47
33540 57.47
33735 57.51
33930 57.51
34125 57.51
34320 57.51
34515 57.51
34710 57.51
34905 57.51
35100 57.51
35295 57.56
35490 57.56
35685 57.56
35880 57.56
36075 57.56
36270 57.56
36465 57.56
36660 57.56
36855 57.56
37050 57.56
37245 57.56
37440 57.6
37635 57.64
37830 57.64
38025 57.77
38220 57.77
38415 57.77
38610 57.77
38805 57.77
39000 57.77
};
\addlegendentry{Alg~\ref{alg:main_alg}- Gaussian}
\addplot [semithick, blue, dotted]
table {%
195 35.67
390 40.22
585 42.77
780 43.44
975 43.44
1170 44.9
1365 46.27
1560 46.77
1755 47.38
1950 47.96
2145 47.96
2340 48.8
2535 49.24
2730 49.25
2925 50.23
3120 50.34
3315 50.68
3510 51.22
3705 51.22
3900 52.15
4095 52.28
4290 52.6
4485 52.7
4680 52.96
4875 52.96
5070 53.41
5265 53.41
5460 53.41
5655 53.41
5850 53.41
6045 53.41
6240 53.84
6435 53.84
6630 53.94
6825 54.17
7020 54.57
7215 54.57
7410 55
7605 55.3
7800 55.3
7995 55.3
8190 55.3
8385 55.57
8580 55.57
8775 55.57
8970 55.57
9165 55.57
9360 55.78
9555 55.78
9750 55.78
9945 55.88
10140 55.88
10335 56.02
10530 56.05
10725 56.08
10920 56.16
11115 56.16
11310 56.28
11505 56.28
11700 56.33
11895 56.47
12090 56.5
12285 56.5
12480 56.5
12675 56.5
12870 56.52
13065 56.52
13260 56.52
13455 56.52
13650 56.52
13845 56.52
14040 56.7
14235 56.7
14430 56.82
14625 56.82
14820 56.82
15015 56.82
15210 56.85
15405 56.96
15600 56.96
15795 56.96
15990 56.96
16185 56.96
16380 56.96
16575 57.22
16770 57.22
16965 57.28
17160 57.28
17355 57.28
17550 57.38
17745 57.51
17940 57.51
18135 57.51
18330 57.51
18525 57.51
18720 57.51
18915 57.51
19110 57.75
19305 57.75
19500 57.75
19695 57.75
19890 57.75
20085 57.75
20280 57.75
20475 57.75
20670 57.75
20865 57.78
21060 57.92
21255 57.97
21450 57.97
21645 57.97
21840 57.97
22035 57.97
22230 57.97
22425 57.98
22620 57.98
22815 58.08
23010 58.08
23205 58.08
23400 58.08
23595 58.08
23790 58.08
23985 58.08
24180 58.08
24375 58.08
24570 58.08
24765 58.08
24960 58.29
25155 58.29
25350 58.29
25545 58.32
25740 58.32
25935 58.32
26130 58.32
26325 58.32
26520 58.32
26715 58.32
26910 58.32
27105 58.32
27300 58.32
27495 58.32
27690 58.32
27885 58.32
28080 58.32
28275 58.32
28470 58.32
28665 58.32
28860 58.32
29055 58.32
29250 58.32
29445 58.34
29640 58.34
29835 58.34
30030 58.34
30225 58.34
30420 58.61
30615 58.61
30810 58.61
31005 58.61
31200 58.61
31395 58.61
31590 58.61
31785 58.61
31980 58.61
32175 58.61
32370 58.61
32565 58.66
32760 58.66
32955 58.74
33150 58.74
33345 58.74
33540 58.75
33735 58.77
33930 58.77
34125 58.84
34320 58.84
34515 58.84
34710 58.84
34905 58.84
35100 58.84
35295 58.84
35490 58.84
35685 58.84
35880 58.87
36075 58.87
36270 58.93
36465 58.93
36660 58.96
36855 58.96
37050 58.98
37245 58.98
37440 58.98
37635 58.98
37830 59.06
38025 59.06
38220 59.18
38415 59.18
38610 59.18
38805 59.18
39000 59.18
};
\addlegendentry{Alg~\ref{alg:main_alg}- Student-t}
\end{axis}

\end{tikzpicture}
	\vspace*{-0.5cm}
	\caption{CIFAR10}
	\end{subfigure}
	\begin{subfigure}[b]{0.48\textwidth}
\begin{tikzpicture}

\begin{axis}[
height=\dualfigureheight,
    width=\dualfigurewidth,
legend cell align={left},
legend style={fill opacity=0.8, draw opacity=1, text opacity=1, at={(0.97,0.03)}, anchor=south east, draw=white!80!black},
tick align=outside,
tick pos=left,
x grid style={white!69.0196078431373!black},
xlabel={\#Iterations},
xmajorgrids,
xmin=-924.3, xmax=24558.3,
xtick style={color=black},
y grid style={white!69.0196078431373!black},
ylabel={Accuracy},
ymajorgrids,
ymin=67.661, ymax=97.779,
ytick style={color=black}
]
\addplot [semithick, black]
table {%
234 69.03
468 76.63
702 81.7
936 84.65
1170 86.32
1404 87.31
1638 88.35
1872 88.77
2106 89.16
2340 89.95
2574 89.98
2808 90.85
3042 90.89
3276 91.29
3510 91.57
3744 91.83
3978 92.01
4212 92.25
4446 92.53
4680 92.7
4914 92.88
5148 92.88
5382 93.04
5616 93.15
5850 93.15
6084 93.17
6318 93.26
6552 93.26
6786 93.49
7020 93.54
7254 93.75
7488 93.75
7722 93.75
7956 93.75
8190 93.75
8424 93.84
8658 93.84
8892 94.03
9126 94.03
9360 94.11
9594 94.2
9828 94.2
10062 94.2
10296 94.31
10530 94.31
10764 94.31
10998 94.31
11232 94.34
11466 94.4
11700 94.44
11934 94.45
12168 94.45
12402 94.45
12636 94.49
12870 94.5
13104 94.55
13338 94.59
13572 94.65
13806 94.69
14040 94.69
14274 94.7
14508 94.7
14742 94.7
14976 94.73
15210 94.73
15444 94.73
15678 94.73
15912 94.73
16146 94.73
16380 94.73
16614 94.73
16848 94.73
17082 94.79
17316 94.79
17550 94.79
17784 94.79
18018 94.79
18252 94.79
18486 94.79
18720 94.79
18954 94.8
19188 94.81
19422 94.81
19656 94.81
19890 94.81
20124 94.81
20358 94.82
20592 94.82
20826 94.82
21060 94.86
21294 94.89
21528 94.89
21762 94.89
21996 94.89
22230 94.89
22464 94.89
22698 94.89
22932 94.89
23166 94.89
23400 94.94
};
\addlegendentry{DPSGD~\cite{abadi2016deep}}
\addplot [semithick, red, dashed]
table {%
234 86.27
468 89.85
702 91.43
936 92.43
1170 92.43
1404 92.43
1638 92.79
1872 92.87
2106 93.13
2340 93.13
2574 93.13
2808 93.48
3042 93.48
3276 93.56
3510 93.82
3744 93.82
3978 94.26
4212 94.26
4446 94.26
4680 94.26
4914 94.26
5148 94.26
5382 94.26
5616 94.26
5850 94.26
6084 94.26
6318 94.26
6552 94.26
6786 94.26
7020 94.26
7254 94.33
7488 94.33
7722 94.33
7956 94.43
8190 94.58
8424 94.58
8658 94.58
8892 94.58
9126 94.58
9360 94.58
9594 94.58
9828 94.59
10062 94.66
10296 94.66
10530 94.79
10764 94.79
10998 95.01
11232 95.12
11466 95.13
11700 95.13
11934 95.13
12168 95.27
12402 95.27
12636 95.27
12870 95.27
13104 95.27
13338 95.36
13572 95.51
13806 95.53
14040 95.53
14274 95.53
14508 95.53
14742 95.58
14976 95.6
15210 95.6
15444 95.6
15678 95.6
15912 95.6
16146 95.75
16380 95.76
16614 95.76
16848 95.8
17082 95.8
17316 95.8
17550 95.8
17784 95.81
18018 95.81
18252 95.81
18486 95.81
18720 95.82
18954 95.82
19188 95.83
19422 95.87
19656 95.89
19890 95.9
20124 95.97
20358 95.97
20592 95.97
20826 95.97
21060 95.97
21294 95.97
21528 95.97
21762 95.97
21996 95.97
22230 96.05
22464 96.05
22698 96.05
22932 96.05
23166 96.05
23400 96.05
};
\addlegendentry{Alg~\ref{alg:main_alg}- Gaussian}
\addplot [semithick, blue, dotted]
table {%
234 87.2
468 90.56
702 92.02
936 92.71
1170 93.35
1404 93.35
1638 93.47
1872 93.86
2106 93.86
2340 93.86
2574 93.86
2808 93.86
3042 94.42
3276 94.42
3510 94.42
3744 94.42
3978 94.42
4212 94.42
4446 94.42
4680 94.42
4914 94.42
5148 94.42
5382 94.42
5616 94.42
5850 94.62
6084 94.74
6318 94.74
6552 94.74
6786 94.85
7020 94.85
7254 94.85
7488 94.94
7722 94.94
7956 94.94
8190 94.94
8424 94.94
8658 94.99
8892 95.02
9126 95.07
9360 95.07
9594 95.07
9828 95.22
10062 95.22
10296 95.42
10530 95.42
10764 95.58
10998 95.58
11232 95.58
11466 95.58
11700 95.58
11934 95.58
12168 95.6
12402 95.7
12636 95.79
12870 95.79
13104 95.79
13338 95.79
13572 95.79
13806 95.79
14040 95.97
14274 95.97
14508 95.97
14742 95.97
14976 95.97
15210 95.97
15444 95.97
15678 96.01
15912 96.01
16146 96.01
16380 96.01
16614 96.01
16848 96.02
17082 96.06
17316 96.13
17550 96.13
17784 96.14
18018 96.14
18252 96.14
18486 96.14
18720 96.15
18954 96.18
19188 96.31
19422 96.31
19656 96.31
19890 96.31
20124 96.31
20358 96.31
20592 96.31
20826 96.31
21060 96.31
21294 96.34
21528 96.34
21762 96.34
21996 96.34
22230 96.34
22464 96.36
22698 96.36
22932 96.38
23166 96.38
23400 96.41
};
\addlegendentry{Alg~\ref{alg:main_alg}- Student-t}
\end{axis}

\end{tikzpicture}
	\vspace*{-0.5cm}
	\caption{MNIST}
	\end{subfigure}
	\caption{Comparing model prediction accuracy for different iterations of training for  constant hyper-parameters (i.e., comparable noise parameters, $\sigma=1.1$ for Gaussian and $\sigma=1$ for Student-t)}
	\label{fig:over_epochs}
\end{figure}

\paragraphb{Training trajectory.}
In Section~\ref{sec:experiments}, we presented results for the final accuracy of our models for different privacy budgets. To demonstrate why our algorithms achieve better results, here we take a look at the accuracy of the models over the training iterations. Figure~\ref{fig:over_epochs} presents the model prediction accuracy for different iterations. When we look at the first iterations of  training, our technique (Algorithm~\ref{alg:main_alg}) yields noticeably higher accuracies compared to  DPSGD for both  MNIST and CIFAR; this allows our algorithm to spend most of the privacy budget for fine-tuning and attaining an overall higher model prediction accuracy.

\paragraphb{Impact of our denoising component.}
To demonstrate the impact of our denoising technique on the overall performance of our framework, we 
 train two models using our framework (Algorithm~\ref{alg:main_alg})  with and without the denoising component. 
Figure~\ref{fig:denoising} compares model accuracies, showing that the denoising component has a substantial impact on improving our overall accuracy (by scaling up privatized gradients that are less noisy).  

\paragraphb{Impact of different noise distributions on model accuracy.}
Figure~\ref{fig:l1_distance} shows the $l1$ distance between  privatized gradient vectors and the corresponding original gradient vectors for different probability distributions, for different \rdp privacy budgets (with $\alpha=2$). 
As we can see  Student-T results in the least  overall noise compared to  other distributions, which leads to models with overall higher prediction accuracies. 

\begin{figure}[h]
\centering
	\begin{minipage}{0.48\textwidth}
	\centering
\begin{tikzpicture}

\begin{axis}[
height=\dualfigureheight,
    width=\dualfigurewidth,
legend cell align={left},
legend style={fill opacity=0.8, draw opacity=1, text opacity=1, at={(0.97,0.03)}, anchor=south east, draw=white!80!black},
tick align=outside,
tick pos=left,
x grid style={white!69.0196078431373!black},
xlabel={\#Iterations},
xmajorgrids,
xmin=-924.3, xmax=24558.3,
xtick style={color=black},
y grid style={white!69.0196078431373!black},
ylabel={Accuracy},
ymajorgrids,
ymin=67.8615, ymax=98.1885,
ytick style={color=black}
]
\addplot [semithick, black]
table {%
234 69.24
468 76.66
702 81.92
936 84.66
1170 86.44
1404 87.28
1638 88.36
1872 88.9
2106 89.07
2340 89.96
2574 90.16
2808 90.79
3042 90.99
3276 91.35
3510 91.62
3744 91.73
3978 92.06
4212 92.21
4446 92.64
4680 92.73
4914 93.06
5148 93.06
5382 93.18
5616 93.22
5850 93.22
6084 93.26
6318 93.45
6552 93.45
6786 93.58
7020 93.6
7254 93.7
7488 93.74
7722 93.8
7956 93.8
8190 93.84
8424 93.84
8658 93.84
8892 94.04
9126 94.04
9360 94.18
9594 94.23
9828 94.23
10062 94.33
10296 94.45
10530 94.45
10764 94.45
10998 94.45
11232 94.59
11466 94.59
11700 94.62
11934 94.62
12168 94.76
12402 94.84
12636 94.84
12870 94.84
13104 94.88
13338 94.99
13572 94.99
13806 95.02
14040 95.02
14274 95.04
14508 95.1
14742 95.1
14976 95.14
15210 95.14
15444 95.19
15678 95.23
15912 95.23
16146 95.27
16380 95.34
16614 95.34
16848 95.39
17082 95.47
17316 95.48
17550 95.6
17784 95.6
18018 95.6
18252 95.6
18486 95.6
18720 95.66
18954 95.66
19188 95.66
19422 95.66
19656 95.66
19890 95.71
20124 95.77
20358 95.81
20592 95.9
20826 95.91
21060 95.91
21294 96.09
21528 96.09
21762 96.09
21996 96.09
22230 96.09
22464 96.11
22698 96.11
22932 96.11
23166 96.16
23400 96.22
};
\addlegendentry{Alg~\ref{alg:main_alg} w.o Denoising}
\addplot [semithick, red, dashed]
table {%
234 86.13
468 87.64
702 91.36
936 92.59
1170 92.85
1404 93.49
1638 93.49
1872 94.01
2106 94.63
2340 94.63
2574 94.7
2808 94.81
3042 94.98
3276 95.24
3510 95.24
3744 95.24
3978 95.45
4212 95.45
4446 95.45
4680 95.45
4914 95.46
5148 95.53
5382 95.53
5616 95.53
5850 95.66
6084 95.66
6318 95.81
6552 95.81
6786 95.81
7020 95.81
7254 95.81
7488 95.81
7722 95.81
7956 95.81
8190 95.81
8424 95.86
8658 95.86
8892 95.86
9126 95.86
9360 95.86
9594 95.86
9828 95.93
10062 95.93
10296 96.11
10530 96.11
10764 96.14
10998 96.15
11232 96.27
11466 96.3
11700 96.3
11934 96.41
12168 96.41
12402 96.41
12636 96.41
12870 96.41
13104 96.41
13338 96.41
13572 96.61
13806 96.61
14040 96.61
14274 96.61
14508 96.62
14742 96.62
14976 96.62
15210 96.62
15444 96.62
15678 96.62
15912 96.62
16146 96.62
16380 96.62
16614 96.62
16848 96.62
17082 96.62
17316 96.67
17550 96.67
17784 96.67
18018 96.67
18252 96.67
18486 96.71
18720 96.71
18954 96.71
19188 96.75
19422 96.75
19656 96.76
19890 96.76
20124 96.8
20358 96.8
20592 96.8
20826 96.8
21060 96.8
21294 96.8
21528 96.8
21762 96.81
21996 96.81
22230 96.81
22464 96.81
22698 96.81
22932 96.81
23166 96.81
23400 96.81
};
\addlegendentry{Alg~\ref{alg:main_alg} w Denoising}
\end{axis}

\end{tikzpicture}
	\vspace*{-0.5cm}
	\caption{Effect of denoising on the learning trajectory}
	\label{fig:denoising}
	\end{minipage}
	\hfill
	\begin{minipage}{0.48\textwidth}
		\centering
\begin{tikzpicture}

\begin{axis}[
height=\dualfigureheight,
    width=\dualfigurewidth,
legend cell align={left},
legend style={fill opacity=0.8, draw opacity=1, text opacity=1, draw=white!80!black},
tick align=outside,
tick pos=left,
x grid style={white!69.0196078431373!black},
xlabel={\(\displaystyle D_{2}\)},
xmajorgrids,
xmin=0.0894432779045333, xmax=0.502694916728926,
xtick style={color=black},
xtick={0,0.1,0.2,0.3,0.4,0.5,0.6},
xticklabels={0.0,0.1,0.2,0.3,0.4,0.5,0.6},
y grid style={white!69.0196078431373!black},
ylabel={L1 Error},
ymajorgrids,
ymin=13.7774226815188, ymax=26.5914977313197,
ytick style={color=black}
]
\addplot [semithick, black]
table {%
0.399 14.3598806383279
0.281 15.9573606771195
0.21 17.5538941043905
0.164 19.1505985088006
0.132 20.7476626461475
0.109 22.3496328380662
};
\addlegendentry{Gaussian}
\addplot [semithick, red, dashed]
table {%
0.372682096034263 14.5806057915553
0.358169004049805 14.7571041038766
0.344526150819377 14.9325117493524
0.331684417484938 15.1050508831286
0.319581407955548 15.2655716167557
0.308160684108435 15.4385360469321
0.297371100086676 15.6101678521394
0.287166220617009 15.7799722111397
0.277503812445658 15.9670512939951
0.268345397967792 16.1313194871155
0.2596558630572 16.3086583538015
0.251403111369167 16.4741615722232
0.2435577592547 16.6452591381211
0.236092865823905 16.8175624578998
0.228983693673284 16.9937203831456
0.222207496480084 17.1597794591679
0.215743329822187 17.3263679671635
0.209571882625694 17.5119412140054
0.20367532679774 17.6755607795625
0.198037182446648 17.8452667371479
0.192642197357189 18.0179064641021
0.187476238678156 18.1907984686314
0.182526195645438 18.36091733957
0.177779891963927 18.5337653365541
0.17322600687129 18.710427980793
0.168854003684561 18.8759938849417
0.164654065518708 19.037671577625
0.160617036689491 19.2241093949313
0.156734369933834 19.3892722146706
0.152998078226475 19.5553459022784
0.149400691215112 19.7333196880075
0.145935215297044 19.9009209907325
0.142595097514948 20.0820826897989
0.139374192383625 20.2466331603211
0.136266731803139 20.4169343134222
0.133267297258997 20.5935481616506
0.130370794753462 20.7680907673809
0.12757243144659 20.9300855368215
0.124867694473297 21.106577941701
0.122252331381331 21.2692315970811
0.119722332256781 21.4403231316882
0.117273913048612 21.6344868310001
0.114903500469706 21.7975026289389
0.112607717941517 21.9563346031294
0.110383372582312 22.1355815523395
0.108227443305642 22.306067136246
};
\addlegendentry{Student-t}
\addplot [semithick, blue, dotted]
table {%
0.483910751327817 16.9913215355229
0.42708422296262 17.6009602492078
0.380062821685845 18.204584698599
0.340696431231714 18.7975982802878
0.307392163250926 19.3834238793526
0.278950629727417 19.9991765075358
0.254455340667863 20.5974353109031
0.233196530476111 21.2086170853117
0.214617761119001 21.809662147235
0.198277880763915 22.4118587125292
0.183823515762948 23.0120158545441
0.170968904254008 23.5928046552957
0.15948092348355 24.1934498225469
0.149167841356937 24.7912822297122
0.139870773266145 25.3973033851207
0.131457126998766 26.0090397745106
};
\addlegendentry{Laplace}
\end{axis}

\end{tikzpicture}
	\vspace*{-0.5cm}
	\caption{The $l1$ distance between privatized gradient vectors and their originals for different \rdp privacy budgets ($D_2$)}
	\label{fig:l1_distance}
	\end{minipage}
\end{figure}

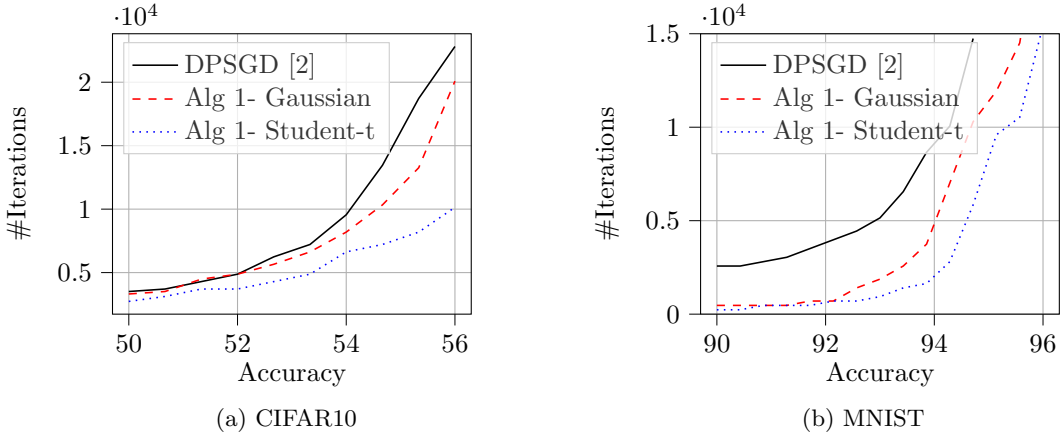
\begin{figure}[h]
\centering
	\begin{subfigure}[b]{0.5\textwidth}
\begin{tikzpicture}

\begin{axis}[
height=\dualfigureheight,
    width=\dualfigurewidth,
legend cell align={left},
legend style={fill opacity=0.8, draw opacity=1, text opacity=1, at={(0.03,0.97)}, anchor=north west, draw=white!80!black},
tick align=outside,
tick pos=left,
x grid style={white!69.0196078431373!black},
xlabel={Accuracy},
xmajorgrids,
xmin=49.7, xmax=56.3,
xtick style={color=black},
y grid style={white!69.0196078431373!black},
ylabel={\#Iterations},
ymajorgrids,
ymin=1725.75, ymax=23819.25,
ytick style={color=black}
]
\addplot [semithick, black]
table {%
50 3510
50.6666666666667 3705
51.3333333333333 4290
52 4875
52.6666666666667 6240
53.3333333333333 7215
54 9555
54.6666666666667 13455
55.3333333333333 18720
56 22815
};
\addlegendentry{DPSGD~\cite{abadi2016deep}}
\addplot [semithick, red, dashed]
table {%
50 3315
50.6666666666667 3510
51.3333333333333 4485
52 4875
52.6666666666667 5655
53.3333333333333 6630
54 8190
54.6666666666667 10335
55.3333333333333 13260
56 20085
};
\addlegendentry{Alg~\ref{alg:main_alg}- Gaussian}
\addplot [semithick, blue, dotted]
table {%
50 2730
50.6666666666667 3120
51.3333333333333 3705
52 3705
52.6666666666667 4290
53.3333333333333 4875
54 6630
54.6666666666667 7215
55.3333333333333 8190
56 10140
};
\addlegendentry{Alg~\ref{alg:main_alg}- Student-t}
\end{axis}

\end{tikzpicture}
	\vspace*{-0.5cm}
	\caption{CIFAR10}
	\end{subfigure}
	\begin{subfigure}[b]{0.48\textwidth}
\begin{tikzpicture}

\begin{axis}[
height=\dualfigureheight,
    width=\dualfigurewidth,
legend cell align={left},
legend style={fill opacity=0.8, draw opacity=1, text opacity=1, at={(0.03,0.97)}, anchor=north west, draw=white!80!black},
tick align=outside,
tick pos=left,
x grid style={white!69.0196078431373!black},
xlabel={Accuracy},
xmajorgrids,
xmin=89.7, xmax=96.3,
xtick style={color=black},
y grid style={white!69.0196078431373!black},
ylabel={\#Iterations},
ymajorgrids,
ymin=0, ymax=15000,
ytick style={color=black}
]
\addplot [semithick, black]
table {%
90 2574
90.4285714285714 2574
90.8571428571429 2808
91.2857142857143 3042
91.7142857142857 3510
92.1428571428571 3978
92.5714285714286 4446
93 5148
93.4285714285714 6552
93.8571428571429 8658
94.2857142857143 10062
94.7142857142857 14742
};
\addlegendentry{DPSGD~\cite{abadi2016deep}}
\addplot [semithick, red, dashed]
table {%
90 468
90.4285714285714 468
90.8571428571429 468
91.2857142857143 468
91.7142857142857 702
92.1428571428571 702
92.5714285714286 1404
93 1872
93.4285714285714 2574
93.8571428571429 3744
94.2857142857143 7020
94.7142857142857 10296
95.1428571428571 11934
95.5714285714286 14508
96 21996
};
\addlegendentry{Alg~\ref{alg:main_alg}- Gaussian}
\addplot [semithick, blue, dotted]
table {%
90 234
90.4285714285714 234
90.8571428571429 468
91.2857142857143 468
91.7142857142857 468
92.1428571428571 702
92.5714285714286 702
93 936
93.4285714285714 1404
93.8571428571429 1638
94.2857142857143 2808
94.7142857142857 5850
95.1428571428571 9594
95.5714285714286 10530
96 15444
};
\addlegendentry{Alg~\ref{alg:main_alg}- Student-t}
\end{axis}

\end{tikzpicture}
	\vspace*{-0.5cm}
	\caption{MNIST}
	\end{subfigure}
	\caption{Comparing the convergence rate for  constant hyper-parameters (i.e., comparable noise parameters, $\sigma=1.1$ for Gaussian, and $\sigma=1$ for Student-t)}
	\label{fig:converge}
\end{figure}

\paragraphb{The convergence rate  of our algorithm.}  The convergence rate of a training algorithm is the number of iterations needed to reach a certain test accuracy; the convergence rate can impact the privacy budget and also the energy consumption. In Figure~\ref{fig:converge}, we compare the convergence rate of our algorithm with DPSGD (for comparable noise parameters, $\sigma=1.1$ for Gaussian, and $\sigma=1$ for Student-t). The results show that Algorithm~\ref{alg:main_alg} can achieve similar accuracies to DPSGD~\cite{abadi2016deep} with in less training iterations,  which results in lower training time, privacy budget and energy consumption.

\section{Conclusions}

Despite their increasing adoption in a wide-range of applications, 
deep learning models are known to leak   information about their training datasets. 
A promising approach to train DNN models with privacy protection is applying differential privacy noise on the gradients during the training process. 
However, existing approaches based on differential privacy result is large degradations in the  utility (prediction accuracy) of the trained models. 
In this work, we design a framework to train DNN models with  differential privacy guarantees while preserving utility significantly better than prior works. 
We specifically introduce two novel techniques to improve the utility-privacy tradeoff. 
First, we encode gradients into a finite vector space; this allows us to obtain privacy bounds for  arbitrary noise distributions applied on the gradients, therefore enabling us to search among different noise distributions for the best privacy-utility tradeoffs. 
Second, we post-process obfuscated gradients, a technique we call denoising, to improve the utility of the trained model without impacting its privacy bounds. 
Our evaluations on two benchmark datasets show that our framework outperforms existing techniques by substantial margins. 
For instance, for the same model accuracy of $96.1\%$ on MNIST, our technique results in a privacy bound  of $\epsilon=3.2$ while the state-of-the-art DPSGD results in  $\epsilon=6$.

\section*{Acknowledgment}
The project has been supported by generous grants from the National Science Foundation (NSF CAREER grant CPS-1739462). Milad Nasr is supported by a Google PhD Fellowship in Security and Privacy.

\bibliographystyle{plain}

\bibliography{refs}
\newpage
\appendix

%
%

\section{Proof of Lemma~\ref{lemma:basic}}

In this section, we present our proof for Lemma~\ref{lemma:basic}. We use Theorem~\ref{lem:lmr} in our proof. 
The overall approach is similar to the proof of the subsampled Gaussian mechanism~\cite{mironov2019r,wang2018subsampled}. 
We limit the \rdp differential privacy analysis only to integer $\alpha$, then we use binomial expansion. Next, we use the addition rule for integrals and reorder them to our final bounds.

\begin{theorem}[Mironov et al.~\cite{mironov2019r}]\label{lem:lmr}
	Let $P$ and $Q$ be two differentiable distributions on $\mathcal{X}$ such that there exists a differentiable
mapping $\nu: \mathcal{X} \mapsto \mathcal{X}$ satisfying $\nu(\nu(x))=x$ and $P(x) = Q(P(\nu(x)))$. Then the following holds for all $\alpha \leq 1$ and $q \in [0,1]$:
\begin{align}
	D_{\alpha}((1-q)P + q Q || Q ) \geq D_{\alpha}( Q||(1-q)P + q Q  )
\end{align}
\end{theorem}

\textbf{Lemma~\ref{lemma:basic}} \textit{Consider one iteration of Algorithm~\ref{alg:main_alg}, and suppose there is only one preselected gradient vector, i.e., $\Psi=\{\vec{\psi}\}$.
Algorithm~\ref{alg:main_alg}  with the sampling rate $q$, and a probability distribution described by its pdf $z(.;\mu)$, where $\mu$ is the mean of the distribution, obeys $(\alpha,\epsilon)-$RDP, for a given $\alpha \in \mathcal{N}\slash \{1\}$;  $\epsilon$ can be computed as follows:}
\begin{align}
    &\epsilon(\alpha; q, \vec{\psi}, z) \leq \\ \nonumber
    & \frac{1}{1-\alpha} \log \sum_{k=0}^{\alpha} \binom{\alpha}{k} q^k (1-q)^{\alpha-k} \prod_{\tau \in  \vec{\psi}} \int_{-\infty}^{\infty} (\frac{z(x;\tau)}{z(x;0)})^k z(x;0) dx
\end{align}
\begin{proof}
	To show our mechanism obeys $\epsilon,\alpha$-\rdp differential privacy, we should show for any two neighbor datasets:
	\begin{align}
		D(\mathcal{M}(D)||\mathcal{M}(D')) \leq \epsilon \\
		D(\mathcal{M}(D')||\mathcal{M}(D)) \leq \epsilon 
	\end{align}
	
We use a  probability distribution $Z(.;\vec{M})$ in our privacy preserving mechanism, where $\vec{M}$ is the mean vector of probability distribution and the covariance is $\Sigma= \sigma I$. Now, for given a subsampling rate $q$, we can rewrite the above mechanism as:
	\begin{align}
		D_\alpha( (1-q)  Z(.;\vec{0}) + qZ(.;\vec{M})||Z(.;\vec{0}))  \leq \epsilon \\
		D_\alpha(Z(.;\vec{0})|| (1-q)  Z(.;\vec{0}) + qZ(.;\vec{M}))\leq \epsilon 
	\end{align}
	
	We should show the validity of the above inequality for every possible mean vector. First, using Theorem~\ref{lem:lmr}, by letting $v(\vec{x})\triangleq \vec{M}-\vec{x} $ we get:
	
			\begin{align}
	 D_\alpha( (1-q)  Z(.;\vec{0}) + qZ(.;\vec{M})||Z(.;\vec{0})) 	\geq D_\alpha(Z(.;\vec{0})|| (1-q)  Z(.;\vec{0}) + qZ(.;\vec{M}))
	\end{align}
		
Now, we only should compute the bound for $ D_\alpha( (1-q)  Z(.;\vec{0}) + qZ(.;\vec{M})||Z(.;\vec{0})) 	$. In lemma~\ref{lemma:basic} we assumed we have only one possible mean vector $M=\vec{\psi}$, so we can compute:
	\begin{align}
		& D_\alpha( (1-q)  Z(.;\vec{0}) + qZ(.;\vec{M})||Z(.;\vec{0})) = D_\alpha( (1-q)  Z(.;\vec{0}) + qZ(.;\vec{\psi})||Z(.;\vec{0})) \\
		& = \frac{1}{1-\alpha} \log \int_{-\infty}^{\infty}\cdots\int_{-\infty}^{\infty} Z(x_1,x_2,\cdots,x_n;\vec{0}) \left( (1-q) + q\frac{ Z(x_1,x_2,\cdots,x_n;\vec{\psi})}{Z(x_1,x_2,\cdots,x_n;\vec{0})} \right)^\alpha dx_1 dx_2 \cdots dx_n
	\end{align}
	Since covariance is diagonal and using the translation invariance of \rdp divergence, we have:
	\begin{align}
		& \frac{1}{1-\alpha} \log \int_{-\infty}^{\infty}\cdots\int_{-\infty}^{\infty} Z(x_1,x_2,\cdots,x_n;\vec{0}) \left( (1-q) + q\frac{ Z(x_1,x_2,\cdots,x_n;\vec{\psi})}{Z(x_1,x_2,\cdots,x_n;\vec{0})} \right)^\alpha dx_1 dx_2 \cdots dx_n \\
		& =\frac{1}{1-\alpha} \log \int_{-\infty}^{\infty}\cdots\int_{-\infty}^{\infty} z(x_1;0)\cdots z(x_n;0) \left( (1-q) + q\frac{ z(x_1;\psi^{(0)})\cdots z(x_n;\psi^{(n)})}{z(x_1;0)\cdots z(x_n;0)} \right)^\alpha dx_1 dx_2 \cdots dx_n
	\end{align}
Given that $\alpha$ is an integer, we can write:
\begin{align}
	& \frac{1}{1-\alpha} \log \int_{-\infty}^{\infty}\cdots\int_{-\infty}^{\infty} z(x_1;0)\cdots z(x_n;0) \left( (1-q) + q\frac{ z(x_1;\psi^{(0)})\cdots z(x_n;\psi^{(n)})}{z(x_1;0)\cdots z(x_n;0)} \right)^\alpha dx_1 dx_2 \cdots dx_n  \\
	&=\frac{1}{1-\alpha} \log \int_{-\infty}^{\infty}\cdots\int_{-\infty}^{\infty} z(x_1;0)\cdots z(x_n;0) \left(\sum_{k=0}^\alpha (1-q)^{\alpha-k} q^{k} (\frac{ z(x_1;\psi^{(0)})\cdots z(x_n;\psi^{(n)})}{z(x_1;0)\cdots z(x_n;0)})^k  \right)  dx_1 dx_2 \cdots dx_n 
\end{align}
Using the addition rule we can rewrite this as:
\begin{align}
	&\frac{1}{1-\alpha} \log \int_{-\infty}^{\infty}\cdots\int_{-\infty}^{\infty} z(x_1;0)\cdots z(x_n;0) \left(\sum_{k=0}^\alpha (1-q)^{\alpha-k} q^{k} (\frac{ z(x_1;\psi^{(0)})\cdots z(x_n;\psi^{(n)})}{z(x_1;0)\cdots z(x_n;0)})^k  \right)  dx_1 dx_2 \cdots dx_n \\
	& = \frac{1}{1-\alpha} \log \sum_{k=0}^\alpha (1-q)^{\alpha-k} q^{k} \int_{-\infty}^{\infty}\cdots\int_{-\infty}^{\infty} z(x_1;0)\cdots z(x_n;0) \left( (\frac{ z(x_1;\psi^{(0)})\cdots z(x_n;\psi^{(n)})}{z(x_1;0)\cdots z(x_n;0)})^k  \right)  dx_1 dx_2 \cdots dx_n \\
	& =   \frac{1}{1-\alpha} \log \sum_{k=0}^{\alpha} \binom{\alpha}{k} q^k (1-q)^{\alpha-k} \prod_{\tau \in  \vec{\psi}} \int_{-\infty}^{\infty} (\frac{z(x;\tau)}{z(x;0)})^k z(x;0) dx
\end{align}

\end{proof}


\end{document}